%% file: main.tex
\documentclass[10pt,twocolumn,letterpaper]{article}

\usepackage[pagenumbers]{cvpr} 

\usepackage{capt-of}
\usepackage[dvipsnames]{xcolor}
\usepackage{xspace}
\usepackage{enumitem}
\usepackage{amsmath,amssymb,amsbsy,amsfonts,dsfont,pifont,bm,bbm,mathrsfs,mathtools,nicefrac}
\usepackage{algorithm,algpseudocode,listings}
\usepackage{booktabs,multirow,adjustbox,diagbox,threeparttable,tabularray}

\newcommand{\method}{LivePhoto\xspace}

\newcommand{\tocite}[1]{{\color{red} [TO CITE]}}

\newcommand{\ve}[1]{\mathbf{#1}} 

\definecolor{cvprblue}{rgb}{0.21,0.49,0.74}
\usepackage[pagebackref,breaklinks,colorlinks,citecolor=cvprblue]{hyperref}
\usepackage{wrapfig}
\usepackage[capitalize]{cleveref}  
\crefname{section}{Sec.}{Secs.}
\Crefname{section}{Section}{Sections}
\crefname{table}{Tab.}{Tabs.}
\Crefname{table}{Table}{Tables}
\crefname{figure}{Fig.}{Figs.}
\Crefname{figure}{Figure}{Figures}
\crefname{equation}{Eq.}{Eqs.}
\Crefname{equation}{Equation}{Equations}
\newcolumntype{x}[1]{>{\centering\arraybackslash}p{#1}}
\newcolumntype{y}[1]{>{\raggedright\arraybackslash}p{#1}}
\newcolumntype{z}[1]{>{\raggedleft\arraybackslash}p{#1}}
\newcommand{\tablestyle}[2]{\setlength{\tabcolsep}{#1}\renewcommand{\arraystretch}{#2}\centering\footnotesize}

\def\paperID{****}
\def\confName{CVPR}
\def\confYear{2024}

\title{\method: Real Image Animation with Text-guided Motion Control}

\author {
    Xi Chen$^{1}$ \quad
    Zhiheng Liu$^{2}$ \quad
    Mengting Chen$^{2}$ \quad
    Yutong Feng$^{2}$ \quad
    \\
    Yu Liu$^{2}$ \quad
    Yujun Shen$^{3}$ \quad
    Hengshuang Zhao$^{1}$ \\[5pt]
    $^{1}$The University of Hong Kong \quad
    $^{2}$Alibaba Group \quad
    $^{3}$Ant Group
}

\begin{document}

\twocolumn[{
\renewcommand\twocolumn[1][]{#1}
\maketitle
\begin{center}
    \vspace{-11pt}
    \includegraphics[width=1.0\linewidth]{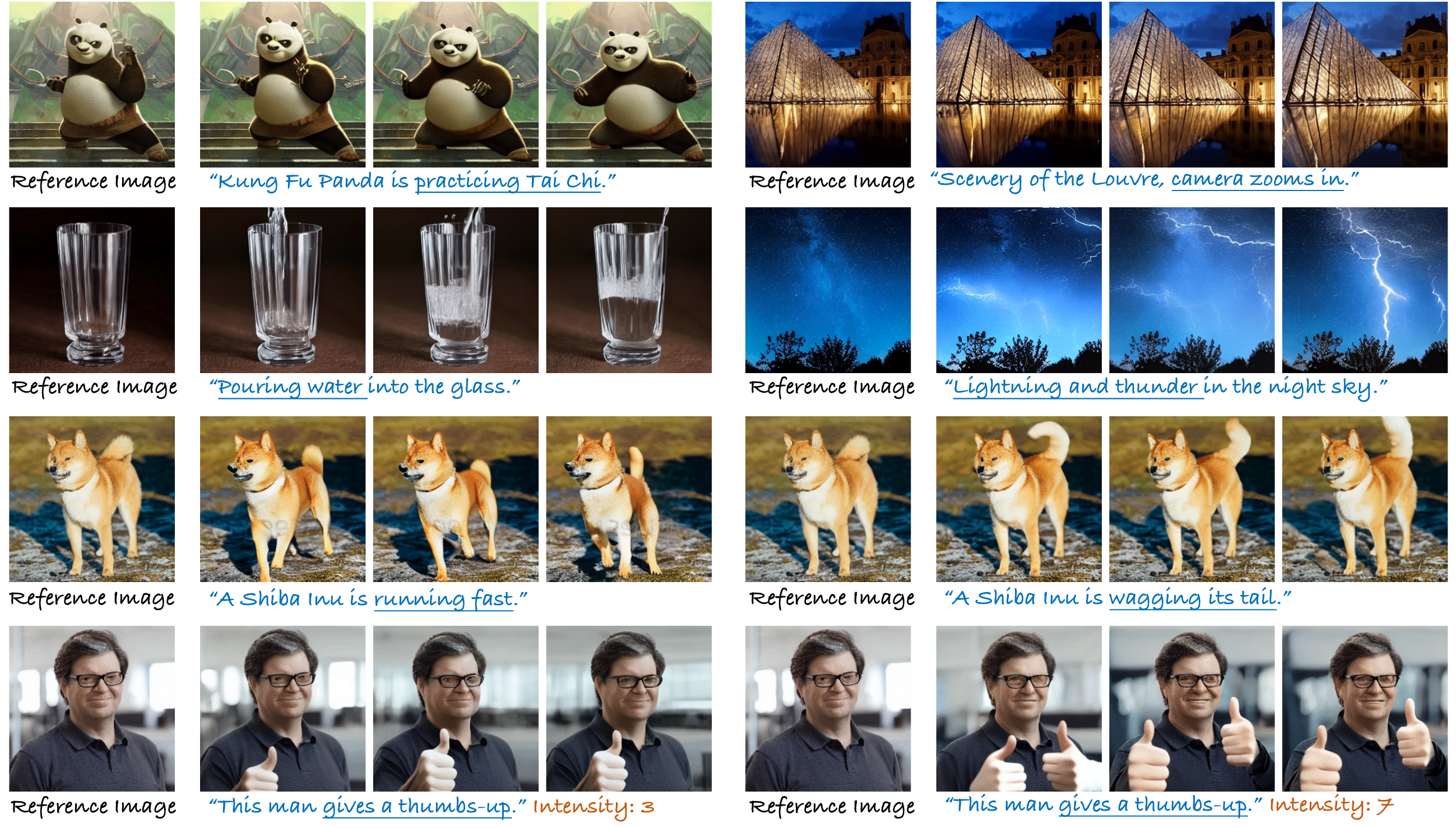}
    \vspace{-18pt}
    \captionsetup{type=figure}
    \caption{%
        \textbf{Zero-shot real image animation with text control.}
        Besides adequately decoding motion descriptions like actions and camera movements~(row 1), \method could also conjure new contents from thin air~(row 2).  
        Meanwhile, \method is highly controllable, supporting users to customize the animation by inputting various texts~(row 3) and adjusting the degree of motion intensity~(row 4).
    }
    \label{fig:teaser}
    \vspace{12pt}
\end{center}
}]

\input{sections/0.abs.tex}
\input{sections/1.intro.tex}
\input{sections/2.related_work.tex}
\input{sections/3.method.tex}
\input{sections/4.exp.tex}
\input{sections/5.conclusion.tex}
\input{sections/6.ref.tex}

\end{document}

%% file: sections/0.abs.tex
\begin{abstract}
\vspace{-5pt}
Despite the recent progress in text-to-video generation, existing studies usually overlook the issue that only spatial contents but not temporal motions in synthesized videos are under the control of text.
Towards such a challenge, this work presents a practical system, named \textbf{\method}, which allows users to animate an image of their interest with text descriptions.
We first establish a strong baseline that helps a well-learned text-to-image generator (\textit{i.e.}, Stable Diffusion) take an image as a further input.
We then equip the improved generator with a motion module for temporal modeling and propose a carefully designed training pipeline to better link texts and motions.
In particular, considering the facts that (1) text can only describe motions roughly (\textit{e.g.}, regardless of the moving speed) and (2) text may include both content and motion descriptions, we introduce a motion intensity estimation module as well as a text re-weighting module to reduce the ambiguity of text-to-motion mapping.
Empirical evidence suggests that our approach is capable of well decoding motion-related textual instructions into videos, such as actions, camera movements, or even conjuring new contents from thin air (\textit{e.g.}, pouring water into an empty glass).
Interestingly, thanks to the proposed intensity learning mechanism, our system offers users an additional control signal (\textit{i.e.}, the motion intensity) besides text for video customization.
The page of this project is \href{https://xavierchen34.github.io/LivePhoto-Page/}{here}.
\end{abstract}
\vspace{-50pt}

%% file: sections/1.intro.tex
\section{Introduction}\label{sec:intro}

Image and video content synthesis has become a burgeoning topic with significant attention and broad real-world applications. Fueled by the diffusion model and extensive training data, image generation has witnessed notable advancements through powerful text-to-image models~\cite{ldm,imagen,xue2023raphael,chen2023pixart} and controllable downstream applications~\cite{controlnet,T2i-adapter,dreambooth, liu2023cones,cones2,imagic,chen2023anydoor}.
In the realm of video generation, a more complex task requiring spatial and temporal modeling, text-to-video has steadily improved~\cite{Make-a-video, Text2video-zero, Align-your-latents, guo2023animatediff, yin2023nuwa}. Various works~\cite{wang2023videocomposer, Tune-a-video, chai2023stablevideo, gen-1, liew2023magicedit} also explore enhancing controllability with sequential inputs like optical flows, motion vectors, depth maps, \textit{etc.}

This work explores utilizing a real image as the initial frame to guide the ``content'' and employ the text to control the ``motion'' of the video. This topic holds promising potential for a wide range of applications, including meme generation, production advertisement, film making, \textit{etc.}
Previous image-to-video methods~\cite{Motion-Conditioned,karras2023dreampose,zhang2023magicavatar,Make-it-move, I2VGen-XL,AnimateDiff-I2V,talesofai} mainly focus on specific subjects like humans or could only animate synthetic images. 
GEN-2~\cite{gen-2} and Pikalabs~\cite{PikaLabs} animate real images with an optional text input, however, an overlooked issue is that the text could only enhance the content but usually fails to control the motions.

Facing this challenge, we propose \method, an image animation framework that truly listens to the text instructions.
We first establish a powerful image-to-video baseline.
The initial step is to equip a text-to-image model~(\textit{i,e., Stable Diffusion}) with the ability to refer to a real image. Specifically, we concatenate the image latent with input noise to provide pixel-level guidance. In addition, a content encoder is employed to extract image patch tokens, which are injected via cross-attention to guide the global identity. During inference, a noise inversion of the reference image is introduced to offer content priors.
Afterward, following the contemporary methods~\cite{Align-your-latents, guo2023animatediff, Tune-a-video}, we freeze stable diffusion models and insert trainable motion layers to model the inter-frame temporal relations. 

Although the text branch is maintained in this strong image-to-video baseline, the model seldom listens to the text instructions. 
The generated videos usually remain nearly static, or sometimes exhibit overly intense movements, deviating from the text.
We identify two key issues for the problem: firstly, the text is not sufficient to describe the desired motion. Phrases like ``shaking the head'' or ``camera zooms in'' lack important information like moving speed or action magnitude.
Thus, a starting frame and a text may correspond to diverse motions with varying intensities. This ambiguity leads to difficulties in linking text and motion. Facing this challenge, we parameterize the motion intensity using a single coefficient, offering a supplementary condition. 
This approach eases the optimization and allows users to adjust motion intensity during inference conveniently.
Another issue arises from the fact that the text contains both content and motion descriptions. The content descriptions translated by stable diffusion may not perfectly align with the reference image, while the image is prioritized for content control. Consequently, when the content descriptions are learned to be suppressed to mitigate conflicts, motion descriptions are simultaneously under-weighted.
To address this concern, we propose text re-weighting, which learns to accentuate the motion descriptions, enabling the text to work compatibly with the image for better motion control.

As shown in \cref{fig:teaser},  equipped with motion intensity guidance and text re-weighting, \method demonstrates impressive abilities for text-guided motion control.
\method is able to deal with real images from versatile domains and subjects, and adequately decodes the motion descriptions like actions and camera movements. Besides, it shows fantastic capacities of conjuring new contents from thin air, like ``pouring water into a glass'' or simulating ``lightning and thunder''. 
In addition, with motion intensity guidance, \method supports users to customize the motion with the desired intensity.

%% file: sections/2.related_work.tex
\section{Related Work}\label{sec:related}

\noindent\textbf{Image animation.}
To realize content controllable video synthesis, image animation takes a reference image as content guidance.
Most of the previous works~\cite{thin-plate-animation, perturbed-masks-animation, sparse-to-dense-animation, first-order-animation,time-flies-animation} depend on another video as a source of motion, transferring the motion to the image with the same subject. 
Other works focus on specific categories like fluide~\cite{fluid-animating,controllable-fluid-animation,eulerian-motion-fluid-animation} or nature objects~\cite{cloud-scene-animating, Generative-image-dynamics}.
Make-it-Move~\cite{Make-it-move} uses text control but it only manipulates simple geometries like cones and cubes.
Recently, human pose transfer methods~\cite{karras2023dreampose, zhang2023magicavatar, wang2023disco, Motion-Conditioned} convert the human images to videos with extra controls like dense poses, depth maps, etc.
VideoComposer~\cite{wang2023videocomposer} could take image and text as controls, however, the text shows limited controllability for the motion and it usually requires more controls like sketches and motion vectors.  
In general, existing work either requires more controls than text or focuses on a specific subject. In this work, we explore constructing a generalizable framework for universal domains and use the most flexible control~(text) to customize the generated video.

\definecolor{flamecolor}{RGB}{235,51,35}
\definecolor{snowflakecolor}{RGB}{47,110,186}
\begin{figure*}[t]
\centering 
\includegraphics[width=1.0\linewidth]{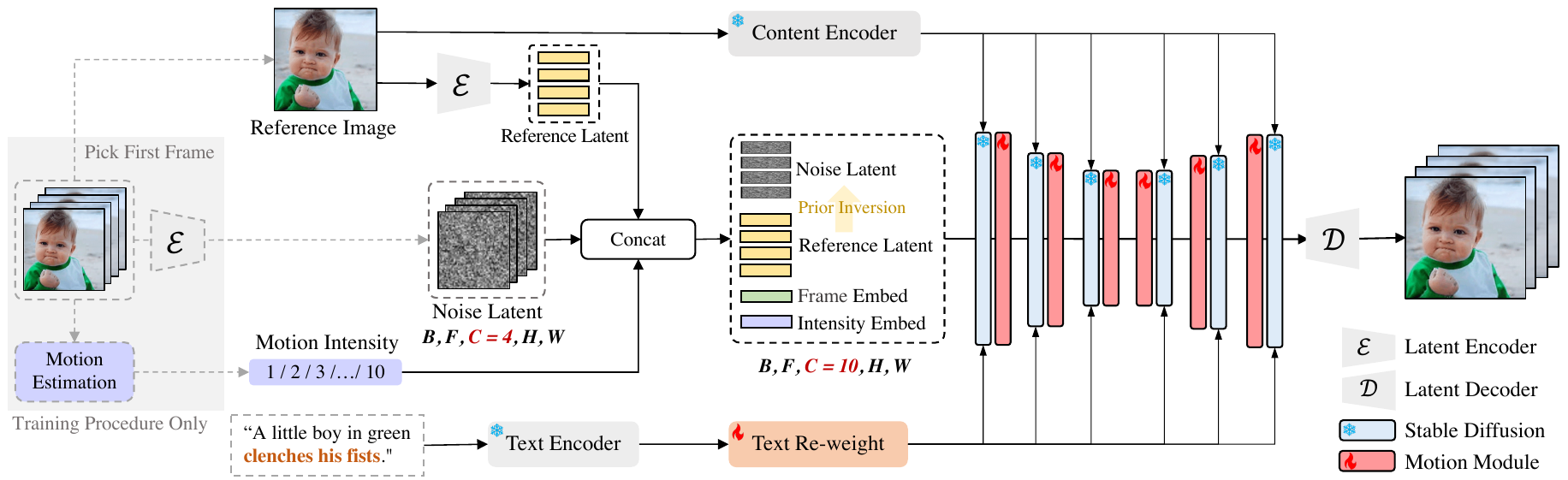} 
\vspace{-15pt}
\caption{
    \textbf{Overall pipeline of \method.} 
    Besides taking the reference image and text as input, \method leverages the motion intensity as a supplementary condition.
    The image and the motion intensity~(from level 1 to 10) are obtained from the ground truth video during training and customized by users during inference.
    The reference latent is first extracted as local content guidance.  We concatenate it with the noise latent, a frame embedding, and the intensity embedding. This 10-channel tensor is fed into the UNet for denoising.  
    During inference, we use the inversion of the reference latent instead of the pure Gaussian to provide content priors.  
    At the top, a content encoder extracts the visual tokens to provide global content guidance. 
    At the bottom, we introduce text re-weighting, which learns to emphasize the motion-related part of the text embedding for better text-motion mapping.
    The visual and textual tokens are injected into the UNet via cross-attention.
    For the UNet, we freeze the pre-trained stable diffusion and insert motion modules to capture the inter-frame relations.
    Symbols of \textbf{\textcolor{flamecolor}{flames}} and \textbf{\textcolor{snowflakecolor}{snowflakes}} denote trainable and frozen parameters respectively.
}
\label{fig:pipeline}
\vspace{-5pt}
\end{figure*}

\noindent\textbf{Text-to-video generation.} 
Assisted by the diffusion model~\cite{ddpm}, the field of text-to-video has progressed rapidly.
Early attempts~\cite{vdm, Make-a-video, yin2023nuwa} train the entire parameters, making the task resource-intensive.
Recently, researchers have turned to leveraging the frozen weights of pre-trained text-to-image models tapping into robust priors.
Tune-A-Video~\cite{Tune-a-video} inflates the text-to-video model and tuning attention modules to construct an inter-frame relationship with a one-shot setting. 
Align-Your-Lantens~\cite{Align-your-latents} inserts newly designed temporal layers into frozen text-to-image models to make video generation.
AnimateDiff~\cite{guo2023animatediff} proposes to freeze the stable diffusion~\cite{ldm} blocks and add learnable motion modules, enabling the model to incorporate with subject-specific LoRAs~\cite{hu2021lora} to make customized generation.
A common issue is that the text could only control the spatial content of the video but exert limited effect for controlling the motions.

%% file: sections/3.method.tex
\section{Method}\label{sec:method}

We first give a brief introduction to the preliminary knowledge for diffusion-based image generation in \cref{sec:preliminaries}. 
Following that, our comprehensive pipeline is outlined in \cref{sec:overall pipeline}. Afterward, \cref{sec:image content} delves into image content guidance to make the model refer to the image. In \cref{sec:motion intensity control} and \cref{sec:text reweighting},  we elaborate on the novel designs of motion intensity guidance and text re-weighting to better align the text conditions with the video motion.

\subsection{Preliminaries} \label{sec:preliminaries}
\noindent\textbf{Text-to-image with diffusion models.} Diffusion models~\cite{ddpm} show promising abilities for both image and video generation. In this work, we opt for the widely used Stable Diffusion~\cite{ldm} as the base model, which adapts the denoising procedure in the latent space with lower computations.
It initially employs VQ-VAE~\cite{VAE} as the latent encoder to transform an image $\ve{x}_0$ into the latent space: $\ve{z}_0 = \mathcal{E}(\ve{x}_0)$. During training, Stable Diffusion transforms the latent into Gaussian noise as follows:

\begin{equation}
    \label{eq:add_noise}
    \ve{z}_t = \sqrt{  \bar{ \alpha_t }} \ve{z}_0 + \sqrt{1- \bar{ \alpha_t }} \ve{\epsilon},
\end{equation}

\noindent where the noise $ \ve{\epsilon}\sim\mathcal{U}([0, 1])$, and  $\bar{ \alpha_t }$ is a cumulative products of the noise coefficient $\alpha_t$ at each step. Afterward, it learns to predict the added noise as:
\begin{equation}
    \label{euq:ldm}
    \mathbb{E}_{\ve{z}, \ve{c},\ve{\epsilon},t}(\| \ve{\epsilon}_{\theta}( \ve{z}_t, \ve{c}, t) - \ve{\epsilon} \|^2_2).
\end{equation}

\noindent $t$ is the diffusion timestep, $\ve{c}$ is the condition of text prompts. During inference, Stable Diffusion is able to recover an image from Gaussian noise step by step by predicting the noise added for each step. The denoising results are fed into a latent decoder to recover the colored images from latent representations as $\ve{\hat x}_0 = \mathcal{D}(\ve{\hat z}_0)$.

\subsection{Overall Pipeline} \label{sec:overall pipeline}
The framework of \method is demonstrated in \cref{fig:pipeline}. 
The model takes a reference image, a text, and the motion intensity as input to synthesize the desired video.  
When the ground truth video is provided during training, the reference image is picked from the first frame, and the motion intensity is estimated from the video. During inference, users could customize the motion intensity or directly use the default level.
\method utilizes a 4-channel tensor of $\mathbf{z}^{B \times F \times C \times H \times W}$ to represent the noise latent of the video, where the dimensions mean batch, frame, channel, height, and width, respectively. 
The reference latent is extracted by VAE encoder~\cite{VAE} to provide local content guidance. 
Meanwhile, the motion intensity is transformed to a 1-channel intensity embedding. 
We concatenate the noise latent, the reference latent, the intensity embedding, and a frame embedding to form a 10-channel tensor for the input of UNet.
At the same time, we use a content encoder to extract the visual tokens of the reference image and inject them via cross-attention.
A text re-weighting module is added after the text encoder~\cite{CLIP}, which learns to assign different weights to each part of the text to accentuate the motion descriptions of the text. 
Following modern text-to-video models~\cite{Align-your-latents,guo2023animatediff}. We freeze the stable diffusion~\cite{ldm} blocks and add learnable motion modules~\cite{guo2023animatediff} at each stage to capture the inter-frame relationships. 

\subsection{Image Content Guidance} \label{sec:image content}
The most essential step is enabling \method to keep the identity of the reference image. Thus, we collect local guidance by concatenating the reference latent at the input. Moreover, we employ a content encoder to extract image tokens for global guidance. Additionally, we introduce the image inversion in the initial noise to offer content priors.

\noindent\textbf{Reference latent.} 
We extract the reference latent and incorporate it at the UNet input to provide pixel-level guidance.  
Simultaneously, a frame embedding is introduced to impart temporal awareness to each frame.
Thus, the first frame could totally trust the reference latent.  Subsequent frames make degenerative references and exhibit distinct behavior.
The frame embedding is represented as a 1-channel map, with values linearly interpolated from zero (first frame) to one (last frame).

\noindent\textbf{Content encoder.} The reference latent effectively guides the initial frames due to their higher pixel similarities. However, as content evolves in subsequent frames, understanding the image and providing high-level guidance becomes crucial. Drawing inspiration from \cite{chen2023anydoor}, we employ a frozen DINOv2~\cite{dinov2} to extract patch tokens from the reference image. We add a learnable linear layer after DINOv2 to project these tokens, which are then injected into the UNet through newly added cross-attention layers.

\noindent\textbf{Prior inversion.} Previous methods~\cite{Tune-a-video,Text2video-zero, meng2021sdedit,talesofai,AnimateDiff-I2V} prove that using an inverted noise of the reference image, rather than a pure Gaussian noise, could effectively provide appearance priors. During inference, we add the inversion of the reference latent $\mathbf{r}_0$ to the noise latent $\mathbf{z}_T^n$ of frame $n$ at the initial denoising step~(T), following \cref{eq:prior}.
\begin{equation}
    \tilde{\ve{z}}_T^n = \alpha^n  \cdot \text{Inv}(\ve{r}_0)+   (1-\alpha^n) \cdot  \ve{z}_T^n, 
    \label{eq:prior}
\end{equation}
where $\alpha^n$ is a descending coefficient from the first frame to the last frame. We set $\alpha^n$ as a linear interpolation from 0.033 to 0.016 by default.

\subsection{Motion Intensity Estimation} \label{sec:motion intensity control}
It is challenging to align the motion coherently with the text. We analyze the core issue is that the text lacks descriptions for the motion speed and magnitude.  Thus, the same text leads to various motion intensities, creating ambiguity in the optimization process.
To address this, we leverage the motion intensity as an additional condition.
We parameterize the motion intensity using a single coefficient. Thus, the users could adjust the intensity conveniently by sliding a bar or directly using the default value.

In our pursuit of parameterizing motion intensity, we experimented with various methods, such as calculating optical flow magnitude, computing mean square error between adjacent frames, and leveraging CLIP/DINO similarity between frames. Ultimately, we found that Structural Similarity (SSIM)~\cite{SSIM} produces results the most aligned with human perceptions.
Concretely, given a training video clip $\mathbf{X}^n$ with n frames, we determine its motion intensity $\mathbf{I}$ by computing the average value for the SSIM~\cite{SSIM} between each adjacent frame as in \cref{eq:intensity} and \cref{eq:ssim}:  
\begin{equation}    
    \mathbf{I}(\mathbf{X}^n) = \frac{1}{n} \sum_{i=0}^{n-2} \text{SSIM}(\mathbf{x}^i, \mathbf{x}^{i+1}).
    \label{eq:intensity}
\end{equation}
\begin{equation}    
    \text{SSIM}(\mathbf{x},\mathbf{y}) = l(\mathbf{x},\mathbf{y})^\alpha \cdot c(\mathbf{x},\mathbf{y})^\beta \cdot s(\mathbf{x},\mathbf{y})^\gamma.
    \label{eq:ssim}
\end{equation}
The structure similarity considers the luminance~($l$), contrast~($c$), and structure~($s$) differences between two images. By default, $\alpha$, $\beta$, and $\gamma$ are set as 1.

We compute the motion intensity on the training data to determine the overall distribution and categorize the values into 10 levels. We create a 1-channel map filled with the level numbers and concatenate it with the input of UNet. During inference, users can utilize level 5 as the default intensity or adjust it between levels 1 to 10. Throughout this paper, unless specified, we use level 5 as the default.

\subsection{Text Re-weighting}  \label{sec:text reweighting}
Another challenge in instructing video motions arises from the fact that the text prompt encompasses both ``content descriptions'' and ``motion descriptions''. The ``content descriptions'', translated by the frozen Stable Diffusion, often fail to perfectly align with the reference images.  When we expect the text prompts to guide the motion, the ``content descriptions'' are inherently accentuated simultaneously. However, as the reference image provides superior content guidance, the effect of the whole text would be suppressed when content conflicts appear. 

\begin{figure}[t]
\centering 
\includegraphics[width=1.0\linewidth]{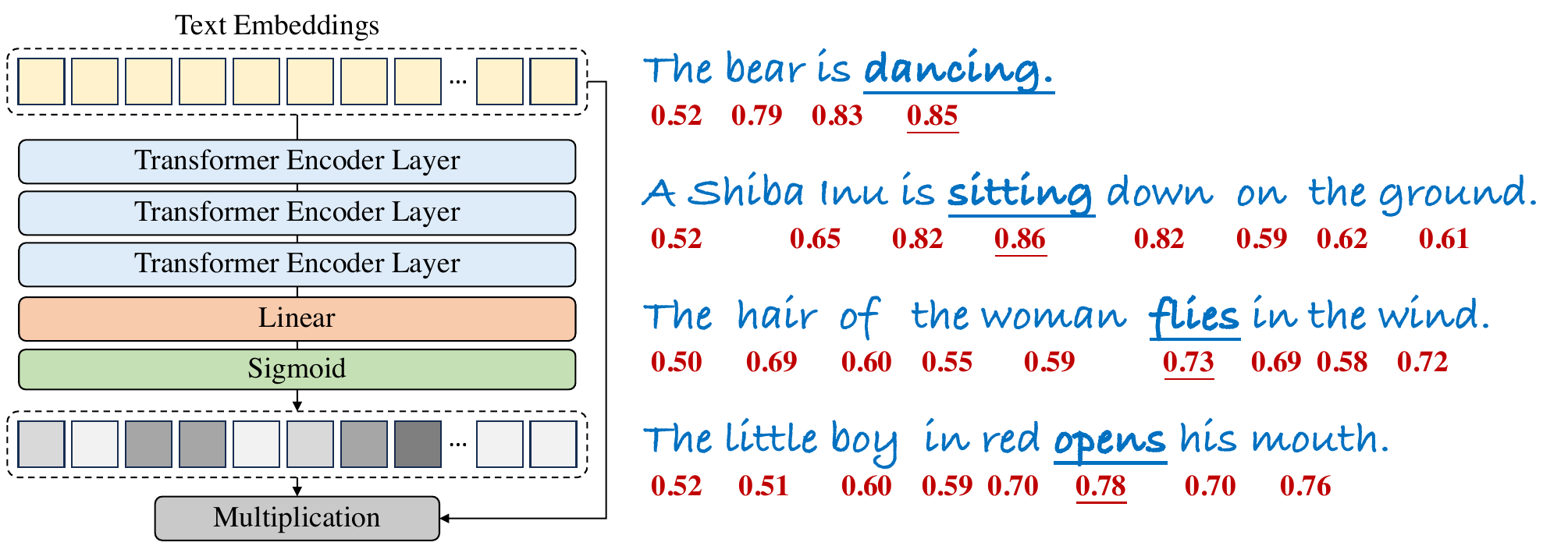} 
\vspace{-18pt}
\caption{%
    \textbf{Demonstrations for text re-weighting.} 
    We use three transformer encoder layers and a frame-specific linear layer to predict the weight for each text token. 
    Examples are given on the right. In cases where multiple tokens correspond to a single word, we calculate the average weight for better visualization. The words with the maximum weight are underlined.
}
\label{fig:reweight_example}
\vspace{-10pt}
\end{figure}

To accentuate the part related to the ``motion descriptions'', we explore manipulating the CLIP text embeddings. Recognizing that directly tuning the text encoder on limited samples might impact generalization, we assign different weights for each embedding without disrupting the CLIP feature space. 
Concretely, we add three trainable transformer layers and a linear projection layer after the CLIP text embeddings. 
Afterward, the predicted weights are normed from 0 to 1 with a sigmoid function. These weights are then multiplied with the corresponding text embeddings, thereby providing guidance that focuses on directing the motions.
The comprehensive structure of the text re-weighting module and actual examples are depicted in \cref{fig:reweight_example}. The numerical results prove that the module successfully learns to emphasize the ``motion descriptions''. This allows signals from images and texts to integrate more effectively, resulting in stronger text-to-motion control.

%% file: sections/4.exp.tex
\section{Experiments}\label{sec:exp}

\subsection{Implementation Details}
\noindent\textbf{Training configurations.} 
We implement \method based on the frozen Stable Diffusion v1.5~\cite{ldm}. The structure of our Motion Module aligns with AnimateDiff~\cite{guo2023animatediff}. Our model is trained on the WebVID~\cite{webvid} dataset employing 8 A100 GPUs. We sample training videos with 16 frames, perform center-cropping, and resize each frame to $256\times256$ pixels. For classifier-free guidance, we utilize a 0.5 probability of dropping the text prompt during training. 
We only use a simple MSE loss to train the model.

\noindent\textbf{Evaluation protocols.} 
We conduct user studies to compare our approach with previous methods and analyze our newly designed modules.
To validate the generalization ability, we gather images from various domains encompassing real images and cartoons including humans, animals, still objects, natural sceneries, \textit{etc.}
For quantitative assessment, we utilize the validation set of WebVID~\cite{webvid}. The first frame and prompt are used as controls to generate videos. We measure the average CLIP similarity~\cite{CLIP} and DINO similarity~\cite{dinov2} between adjacent frames to evaluate the frame consistency following previous works~\cite{wang2023videocomposer,gen-1}.

\subsection{Ablation Studies}
In this section, we thoroughly analyze each of our proposed modules to substantiate their effectiveness.  We first analyze how to add content guidance with the reference image, which is an essential part of our framework.  Following that, we delve into the specifics of our newly introduced motion intensity guidance and text re-weighting.

\begin{figure}[t]
\centering 
\includegraphics[width=1.0\linewidth]{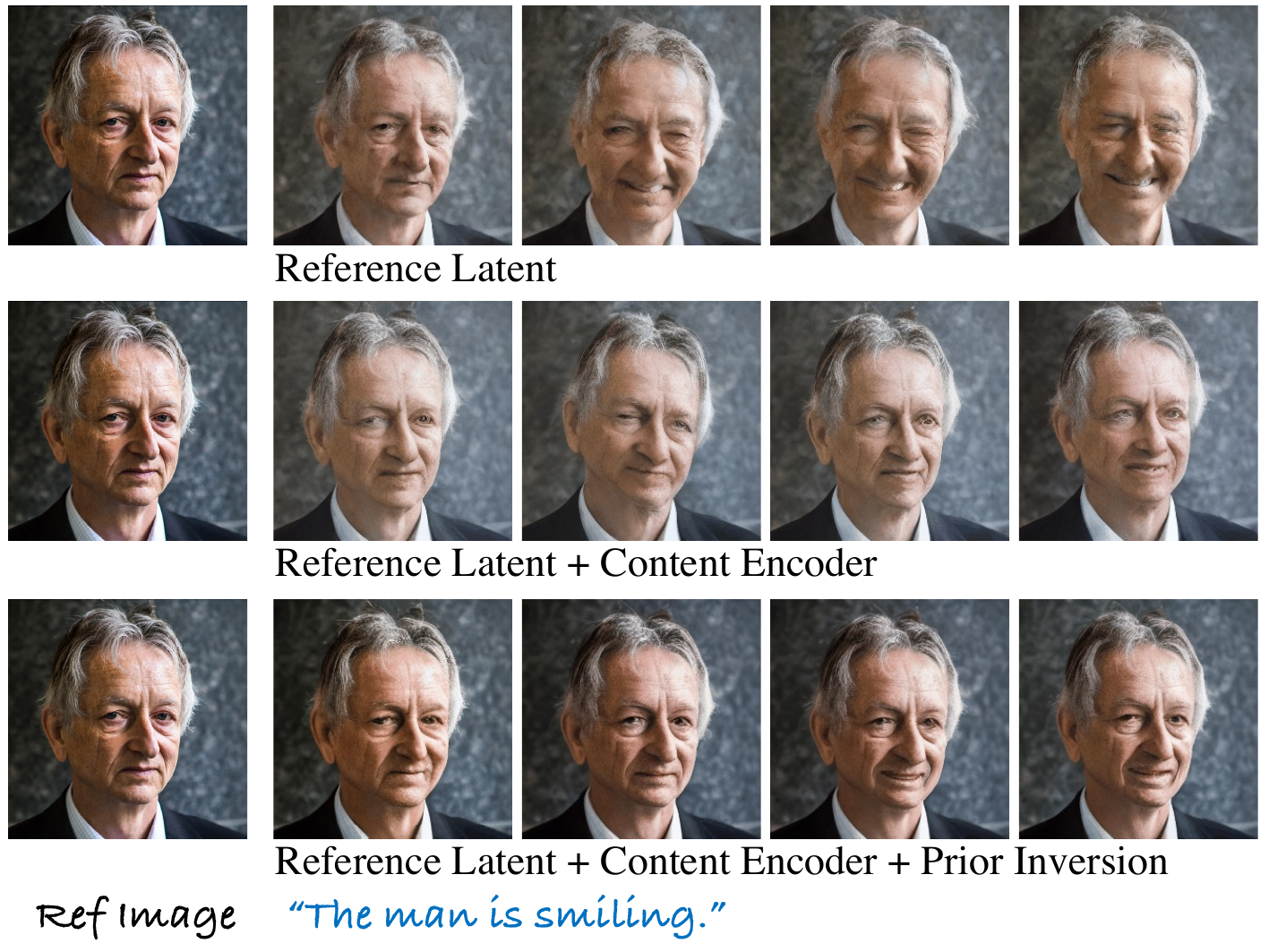} 
\vspace{-18pt}
\caption{%
    \textbf{Ablations for the image content guidance.} 
    Only concatenating the reference latent with the model input meets challenges in preserving the identity. 
    The content encoder and prior inversion gradually enhance the performance.
}
\label{fig:imagecontentguidance}
\vspace{-10pt}
\end{figure}

\noindent\textbf{Image content guidance.}  As introduced in \cref{sec:overall pipeline}, we concatenate the reference latent with the input as the pixel-wise guidance and use a content encoder to provide the holistic identity information. Besides, the prior inversion further assists the generation of details. 
In \cref{fig:imagecontentguidance}, we illustrate the step-by-step integration of these elements. In row~1, the reference latent could only keep the identity for the starting frames as the contents are similar to the reference image. After adding the content encoder in row~2, the identity for the subsequent frames could be better preserved but the generation quality for the details is not satisfactory. With the inclusion of prior inversion, the overall quality sees further improvement.  The quantitative results in \cref{tab:content} consistently confirm the effectiveness of each module. These three strategies serve as the core of our strong baseline for real image animation.

\begin{table}[t]
\caption{%
    \textbf{Quatitative analysis for image content guidance.} 
    We assess frame consistency using DINO and CLIP scores. 
    The content encoder and prior inversion bring steady improvements. }
\label{tab:content}
\vspace{-7pt}
\centering\footnotesize
\setlength{\tabcolsep}{9pt}
\begin{tabular}{lcc}
\toprule
Method  & DINO Score~($\uparrow$) & CLIP Score~($\uparrow$)  \\
\midrule
Reference Latent        & 82.3  &  91.7  \\
+ Content Encoder      & 85.9  &  93.2  \\
++ Prior Inversion   & \textbf{90.8}  & \textbf{95.2}   \\
\bottomrule
\end{tabular}
\vspace{-10pt}
\end{table}

\begin{figure*}[t]
\centering 
\includegraphics[width=1.0\linewidth]{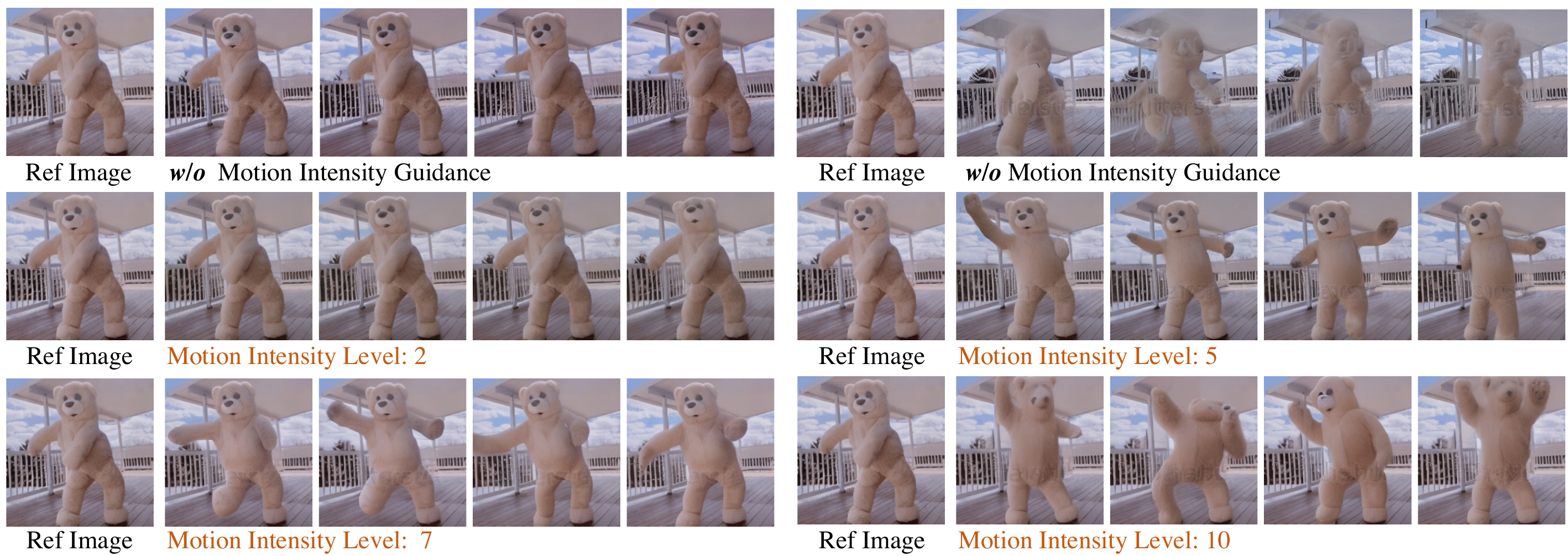} 
\vspace{-18pt}
\caption{%
    \textbf{Illustrations of motion intensity guidance.} The prompt is \textbf{\textcolor{snowflakecolor}{``The bear is dancing''.}}
    Without intensity guidance, the generated video tends to either keep still or quickly become blurry. 
    With the option to set varying intensity levels, users can finely control the motion range and speed. 
    It should be noted that excessively high intensity levels might induce motion blur, as observed in the last case.
}
\label{fig:intensity}
\vspace{-10pt}
\end{figure*}

\begin{figure}[t]
\centering 
\includegraphics[width=1.0\linewidth]{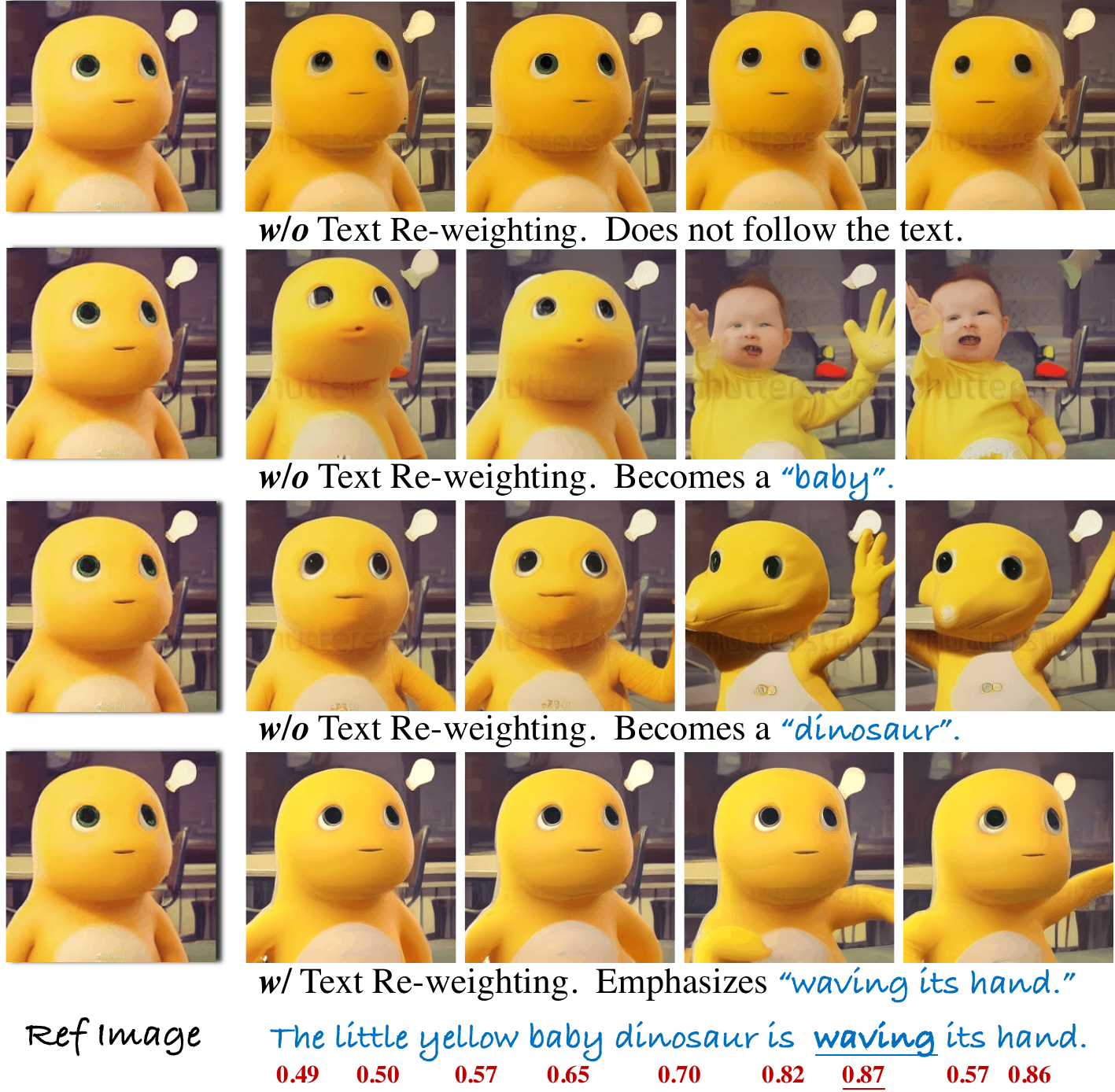} 
\vspace{-18pt}
\caption{%
    \textbf{Ablation for text re-weighting.} 
    Without re-weighting, the model tends to either disregard the text entirely or fixate on content-related descriptions like ``baby dinosaur''.
    When re-weighting is applied, content descriptions are suppressed while motion-related details like ``waving its hand'' gain emphasis.
    The predicted weights of text re-weighting are marked at the bottom.
}
\label{fig:ab_reweighting}
\vspace{-10pt}
\end{figure}

\noindent\textbf{Motion intensity guidance.} As introduced in \cref{sec:motion intensity control}, we parameterize the motion intensity as a coefficient, and use it to indicate the motion speed and ranges. We carry out ablation studies in \cref{fig:intensity}. The absence of motion intensity guidance often leads to static or erratic video outputs, as depicted in the first row. 
However, with the introduction of intensity guidance, the subsequent rows display varying motion levels, allowing for the production of high-quality videos with different motion ranges. Notably, lower levels like level 2 generate almost static videos, while higher levels like 10 occasionally produce overly vigorous motions. Users could directly use the default value~(level~5) or tailor the intensity according to specific preferences.

\noindent\textbf{Text re-weighting.} In \cref{fig:ab_reweighting}, we demonstrate the efficacy of text re-weighting.  In the given examples, the content description ``baby dinosaur'' would conflict with the reference image. In the first three rows, without the assistance of re-weighting, the frozen Stabel Diffusion tends to synthesize the content through its understanding of the text. Thus, the produced video tends to ignore the text and follow the reference image as in row 1.  In other cases, it has risks of becoming a ``baby''~(row 2) or a ``dinosaur''~(row 3).  As visualized in the bottom of \cref{fig:ab_reweighting}, text re-weighting elevates emphasis on motion descriptions like ``waving its hand''. This approach enables our model to faithfully follow text-based instructions for motion details while upholding image-consistent content with the reference image.

The quantitative results are listed in \cref{tab:newmodule}. The motion intensity guidance and text re-weighting both contribute to the frame consistency.

\begin{table}[t]
\caption{%
    \textbf{Quatitative analysis for novel modules.} 
    Frame consistency is measured by DINO and CLIP scores.
    Motion intensity guidance and text re-weighting both make contributions. 
}
\label{tab:newmodule}
\vspace{-7pt}
\centering\footnotesize
\setlength{\tabcolsep}{9pt}
\begin{tabular}{lcc}
\toprule
Method   & DINO Score~($\uparrow$)  & CLIP Score~($\uparrow$)\\
\midrule
\method       & \textbf{90.8}  & \textbf{95.2}   \\
\textit{w/o} Motion Intensity    & 90.3  & 94.8   \\
\textit{w/o} Text Re-weighting   & 90.1  & 93.9   \\
\bottomrule
\end{tabular}
\vspace{-10pt}
\end{table}

\subsection{Comparisons with Existing Alternatives} 
We compare \method with other works that support image animation with text control. 
VideoComposer~\cite{wang2023videocomposer} is a strong compositional generator covering various conditions including image and text.
GEN-2~\cite{gen-2} and Pikalabs~\cite{PikaLabs} are famous products that support image and text input. 
I2VGEN-XL~\cite{I2VGen-XL}, AnimateDiff-I2V~\cite{AnimateDiff-I2V}, Talesofai~\cite{talesofai} are open-source projects claiming similar abilities.

\begin{figure*}[t]
\centering 
\includegraphics[width=1.0\linewidth]{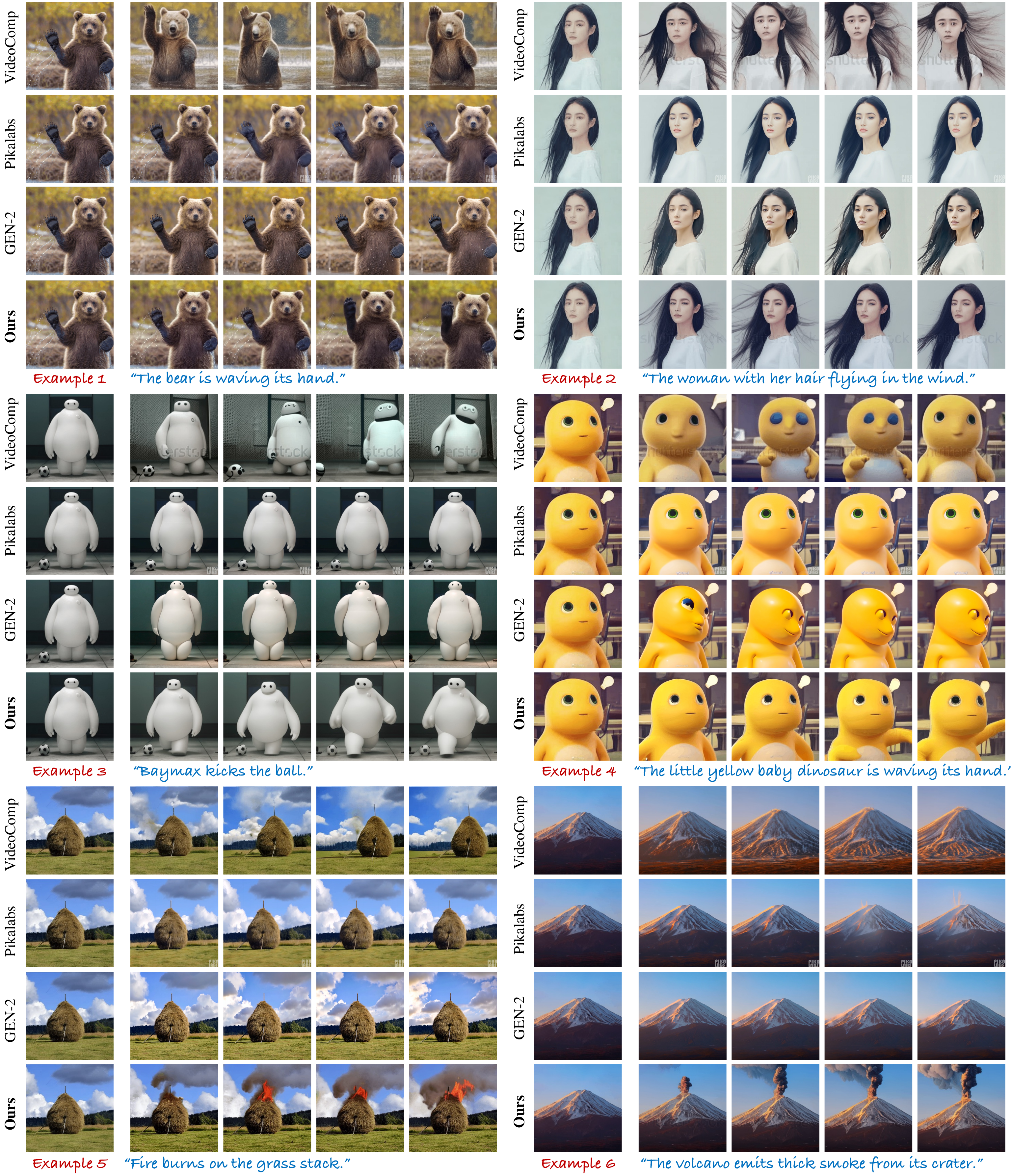} 
\vspace{-18pt}
\caption{%
    \textbf{Comparison results with other methods.} We compare our \method with VideoComposer~\cite{wang2023videocomposer}, Pikalabs~\cite{PikaLabs}, and GEN-2~\cite{gen-2}.
    We select representative cases covering animal, human, cartoon, and natural scenery. 
    To ensure a fair evaluation, we executed each method 8 times, presenting the most optimal outcomes for comparison. 
    In each example, the reference image is displayed on the left, accompanied by the text prompt indicated at the bottom.
}
\label{fig:compare}
\vspace{-10pt}
\end{figure*}

\noindent\textbf{Qualitative analysis.} 
In \cref{fig:compare}, we compare \method with VideoComposer~\cite{wang2023videocomposer}, Pikalabs~\cite{PikaLabs}, and GEN-2~\cite{gen-2}
with representative examples. The selected examples cover animals, humans, cartoons, and natural scenarios. To reduce the randomness, we ran each method 8 times to select the best result for more fair comparisons.
VideoComposer demonstrates proficiency in creating videos with significant motion. However, as not specifically designed for photo animation, the identity-keeping ability is not satisfactory. The identities of the reference images are lost, especially for less commonly seen subjects. Additionally, it shows a lack of adherence to the provided text instructions.
Pikalabs~\cite{PikaLabs} and GEN-2~\cite{gen-2} produce high-quality videos. However, as a trade-off, the generated videos own limited motion ranges. Although they support text as supplementary, the text descriptions seldom work. The motions are generally estimated from the content of the reference image. 

In contrast, \method adeptly preserves the identity of the reference image and generates consistent motions with the text instructions. It performs admirably across various domains, encompassing animals, humans, cartoon characters, and natural sceneries. It not only animates specific actions~(examples~1-4) but also conjures new effects from thin air~(examples~5-6).

\begin{figure}[t]
\centering 
\includegraphics[width=1.0\linewidth]{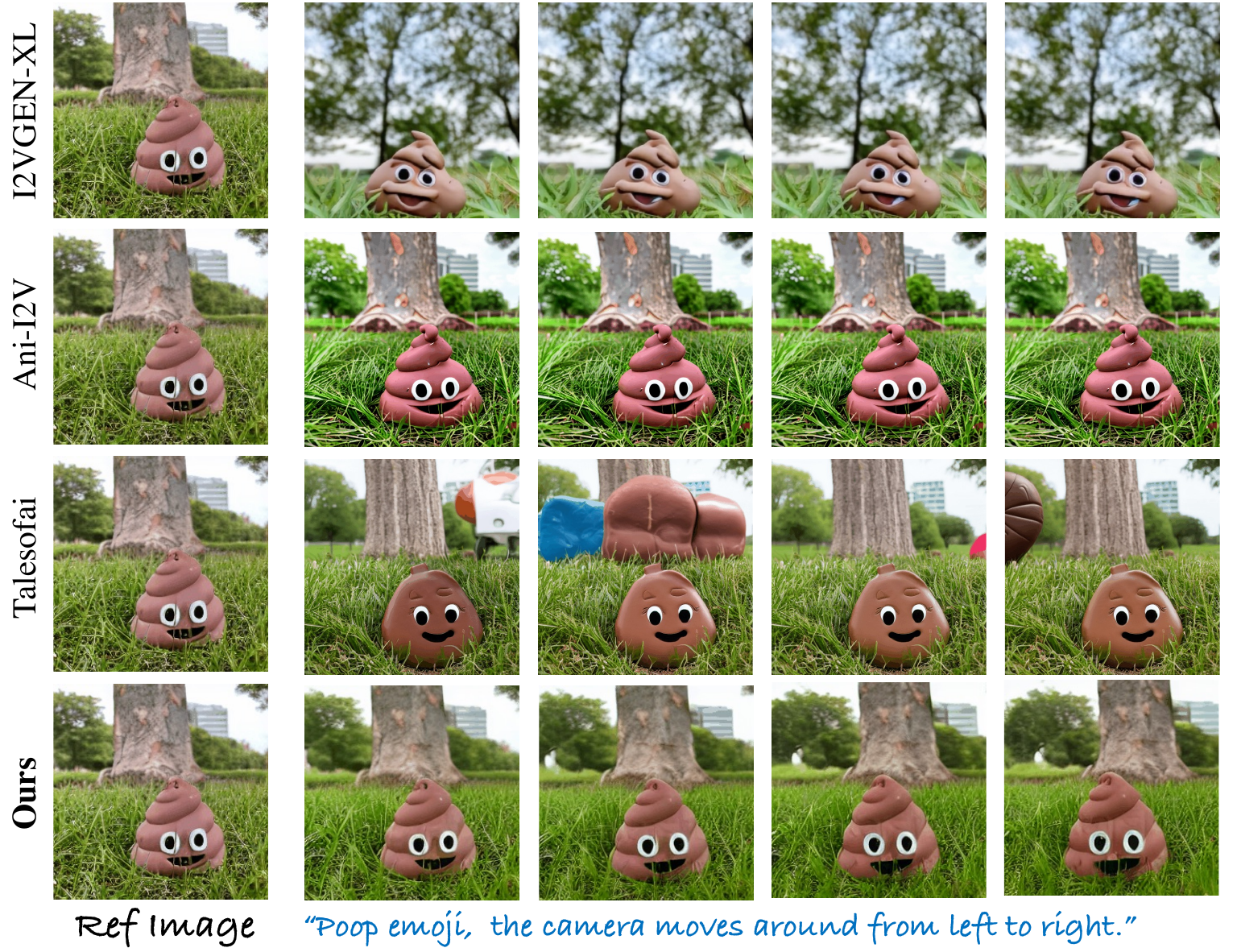} 
\vspace{-18pt}
\caption{%
    \textbf{Comparisons with open-sourced projects.} 
    I2VGEN-XL~\cite{I2VGen-XL}, AnimateDiff-I2V~\cite{AnimateDiff-I2V}, and Talesofai~\cite{talesofai} also support animating an image with text. However, I2VGEN-XL only generates ``relevant'' content with the reference image.  The produced videos of AnimateDiff-I2V rarely move. Talesofai could not keep the identity for real photos.
}
\label{fig:project}
\vspace{-15pt}
\end{figure}

We also compare \method with open-sourced project in \cref{fig:project}. I2VGEN-XL~\cite{I2VGen-XL} does not set the reference image as the first frame but generates videos with similar semantics.  AnimateDiff-I2V~\cite{AnimateDiff-I2V} and Talsofai~\cite{talesofai} are extensions of AnimateDiff~\cite{guo2023animatediff}. However, the former produces quasi-static videos. The latter fails to keep the image identity unless using SD-generated images with the same prompt and corresponding LoRA~\cite{hu2021lora}.

\begin{table}[t]
\caption{%
    \textbf{Results of user study.}
    We let annotators rate from four perspectives: Image consistency~($\ve{C}_{\text{image}}$) evaluates the capability to maintain the identity of the reference image.  Text consistency~($\ve{C}_{\text{text}}$) measures the adherence to the textual descriptions in directing motion.  Content quality~($\ve{Q}_{\text{cont}}$) focuses on the inter-frame coherence and resolutions. Motion quality~($\ve{Q}_{\text{mot}}$) evaluates appropriateness of motions. 
}
\label{tab:userstudy}
\vspace{-7pt}
\centering\footnotesize
\setlength{\tabcolsep}{9pt}
\tablestyle{3.5pt}{1.05}
\begin{tabular}{lcccc}
\toprule
Method  & $\ve{C}_{\text{image}}~(\uparrow)$ & $\ve{C}_{\text{text}}~(\uparrow)$ & $\ve{Q}_{\text{cont}}~(\uparrow)$ & $\ve{Q}_{\text{mot}}~(\uparrow)$ \\
\midrule
VideoComposr~\cite{wang2023videocomposer}    & 2.8 & 3.5 & 3.6 & 3.6  \\
Pikalabs~\cite{PikaLabs}  & \textbf{3.9} & 2.7 & 4.6 & 3.1  \\
GEN-2~\cite{gen-2}      & 3.7 & 2.5 & \textbf{4.8} & 3.3  \\
\midrule
\method    & 3.6 & \textbf{4.7} & 3.7 & \textbf{3.9}  \\
\textit{w/o} text re-weighting   & 3.5 & 3.3 & 3.6 & 3.8  \\
\textit{w/o} intensity guidance   & 3.4 & 2.5 & 3.4 & 3.5  \\
\bottomrule
\end{tabular}
\vspace{-10pt}
\end{table}

\noindent\textbf{User studies.} 
Metrics like DINO/CLIP scores have limitations in thoroughly evaluating the model, thus, we carry out user studies. We ask the annotators to rate the generated videos from 4 perspectives: Image consistency evaluates the identity-keeping ability of the reference image. Text consistency measures whether the motion follows the text descriptions. Content quality considers the general quality of videos like the smoothness, the resolution, \textit{etc.} Motion quality assesses the reasonableness of generated motion, encompassing aspects such as speed and deformation.

We construct a benchmark with five tracks: humans, animals, cartoon characters, still objects, and natural sceneries. We collect 10 reference images per track and manually write 2 prompts per image. Considering the variations that commonly exist in video generation, each method is required to predict 8 results. Thus, we get 800 samples for each method.  We first ask 4 annotators to pick the best ones out of 8 predictions according to the aforementioned four perspectives. Then, we ask 10 annotators to further rate the filtered samples. 
As the projects~\cite{I2VGen-XL,AnimateDiff-I2V,talesofai} demonstrates evidently inferior results, we only compare \method with VideoComposer~\cite{wang2023videocomposer}, GEN-2~\cite{gen-2}, and Pikalabs~\cite{PikaLabs}. 

Results in \cref{tab:userstudy} demonstrate that GEN-2\cite{gen-2} and Pikalabs own slightly better image consistency because their generated video seldom moves. \method shows significantly better text consistency and motion quality compared with other works. We admit that GEN-2 and Pikalabs own superior smoothness and resolution. We infer that they might collect much better training data and leverage super-resolution networks as post-processing. However, as an academic method, \method shows distinguishing advantages over mature products in certain aspects. We have reasons to believe its potential for future applications.
\vspace{-8pt}

%% file: sections/5.conclusion.tex
\section{Limitations}\label{sec:limitations}
\vspace{-6pt}
\method is implemented on SD-1.5 with $256\times256$ output considering the training cost.
We believe that with higher resolution and stronger models like SD-XL~\cite{podell2023sdxl}, the overall performance could be further improved significantly.
\vspace{-6pt}

\section{Conclusion}\label{sec:conclusion}
\vspace{-6pt}
\noindent We introduce \method, a novel framework for photo animation with text control. 
We propose a strong baseline that gathers the image content guidance from the given image and utilizes motion intensity as a supplementary to better capture the desired motions. 
Besides, we propose text re-weighting to accentuate the motion descriptions. 
The whole pipeline illustrates impressive performance for generalized domains and instructions. 

%% file: sections/6.ref.tex
{
\small
\bibliographystyle{ieeenat_fullname}
\bibliography{ref.bib}
}

%% file: main.bbl
\begin{thebibliography}{51}
\providecommand{\natexlab}[1]{#1}
\providecommand{\url}[1]{\texttt{#1}}
\expandafter\ifx\csname urlstyle\endcsname\relax
  \providecommand{\doi}[1]{doi: #1}\else
  \providecommand{\doi}{doi: \begingroup \urlstyle{rm}\Url}\fi

\bibitem[Bain et~al.(2021)Bain, Nagrani, Varol, and Zisserman]{webvid}
Max Bain, Arsha Nagrani, G{\"u}l Varol, and Andrew Zisserman.
\newblock Frozen in time: A joint video and image encoder for end-to-end retrieval.
\newblock In \emph{ICCV}, 2021.

\bibitem[Blattmann et~al.(2023)Blattmann, Rombach, Ling, Dockhorn, Kim, Fidler, and Kreis]{Align-your-latents}
Andreas Blattmann, Robin Rombach, Huan Ling, Tim Dockhorn, Seung~Wook Kim, Sanja Fidler, and Karsten Kreis.
\newblock Align your latents: High-resolution video synthesis with latent diffusion models.
\newblock In \emph{CVPR}, 2023.

\bibitem[Chai et~al.(2023)Chai, Guo, Wang, and Lu]{chai2023stablevideo}
Wenhao Chai, Xun Guo, Gaoang Wang, and Yan Lu.
\newblock Stablevideo: Text-driven consistency-aware diffusion video editing.
\newblock In \emph{ICCV}, 2023.

\bibitem[Chen et~al.(2023{\natexlab{a}})Chen, Yu, Ge, Yao, Xie, Wu, Wang, Kwok, Luo, Lu, et~al.]{chen2023pixart}
Junsong Chen, Jincheng Yu, Chongjian Ge, Lewei Yao, Enze Xie, Yue Wu, Zhongdao Wang, James Kwok, Ping Luo, Huchuan Lu, et~al.
\newblock Pixart: Fast training of diffusion transformer for photorealistic text-to-image synthesis.
\newblock \emph{arXiv:2310.00426}, 2023{\natexlab{a}}.

\bibitem[Chen et~al.(2023{\natexlab{b}})Chen, Lin, Tseng, Lin, and Yang]{Motion-Conditioned}
Tsai-Shien Chen, Chieh~Hubert Lin, Hung-Yu Tseng, Tsung-Yi Lin, and Ming-Hsuan Yang.
\newblock Motion-conditioned diffusion model for controllable video synthesis.
\newblock \emph{arXiv:2304.14404}, 2023{\natexlab{b}}.

\bibitem[Chen et~al.(2023{\natexlab{c}})Chen, Huang, Liu, Shen, Zhao, and Zhao]{chen2023anydoor}
Xi Chen, Lianghua Huang, Yu Liu, Yujun Shen, Deli Zhao, and Hengshuang Zhao.
\newblock Anydoor: Zero-shot object-level image customization.
\newblock \emph{arXiv:2307.09481}, 2023{\natexlab{c}}.

\bibitem[Cheng et~al.(2020)Cheng, Chen, and Chiu]{time-flies-animation}
Chia-Chi Cheng, Hung-Yu Chen, and Wei-Chen Chiu.
\newblock Time flies: Animating a still image with time-lapse video as reference.
\newblock In \emph{CVPR}, 2020.

\bibitem[Esser et~al.(2023)Esser, Chiu, Atighehchian, Granskog, and Germanidis]{gen-1}
Patrick Esser, Johnathan Chiu, Parmida Atighehchian, Jonathan Granskog, and Anastasis Germanidis.
\newblock Structure and content-guided video synthesis with diffusion models.
\newblock In \emph{ICCV}, 2023.

\bibitem[group(2023)]{I2VGen-XL}
Alibaba group.
\newblock I2vgen-xl.
\newblock \url{https://modelscope.cn/models/damo/Video-to-Video/summary}, 2023.

\bibitem[Guo et~al.(2023)Guo, Yang, Rao, Wang, Qiao, Lin, and Dai]{guo2023animatediff}
Yuwei Guo, Ceyuan Yang, Anyi Rao, Yaohui Wang, Yu Qiao, Dahua Lin, and Bo Dai.
\newblock Animatediff: Animate your personalized text-to-image diffusion models without specific tuning.
\newblock \emph{arXiv:2307.04725}, 2023.

\bibitem[Ho et~al.(2020)Ho, Jain, and Abbeel]{ddpm}
Jonathan Ho, Ajay Jain, and Pieter Abbeel.
\newblock Denoising diffusion probabilistic models.
\newblock \emph{NeurIPS}, 2020.

\bibitem[Ho et~al.(2022)Ho, Salimans, Gritsenko, Chan, Norouzi, and Fleet]{vdm}
Jonathan Ho, Tim Salimans, Alexey Gritsenko, William Chan, Mohammad Norouzi, and David~J Fleet.
\newblock Video diffusion models.
\newblock \emph{arXiv:2204.03458}, 2022.

\bibitem[Holynski et~al.(2021)Holynski, Curless, Seitz, and Szeliski]{eulerian-motion-fluid-animation}
Aleksander Holynski, Brian~L Curless, Steven~M Seitz, and Richard Szeliski.
\newblock Animating pictures with eulerian motion fields.
\newblock In \emph{CVPR}, 2021.

\bibitem[Hu et~al.(2021)Hu, Shen, Wallis, Allen-Zhu, Li, Wang, Wang, and Chen]{hu2021lora}
Edward~J Hu, Yelong Shen, Phillip Wallis, Zeyuan Allen-Zhu, Yuanzhi Li, Shean Wang, Lu Wang, and Weizhu Chen.
\newblock Lora: Low-rank adaptation of large language models.
\newblock \emph{arXiv:2106.09685}, 2021.

\bibitem[Hu et~al.(2022)Hu, Luo, and Chen]{Make-it-move}
Yaosi Hu, Chong Luo, and Zhenzhong Chen.
\newblock Make it move: controllable image-to-video generation with text descriptions.
\newblock In \emph{CVPR}, 2022.

\bibitem[Jhou and Cheng(2015)]{cloud-scene-animating}
Wei-Cih Jhou and Wen-Huang Cheng.
\newblock Animating still landscape photographs through cloud motion creation.
\newblock \emph{TMM}, 2015.

\bibitem[Karras et~al.(2023)Karras, Holynski, Wang, and Kemelmacher-Shlizerman]{karras2023dreampose}
Johanna Karras, Aleksander Holynski, Ting-Chun Wang, and Ira Kemelmacher-Shlizerman.
\newblock Dreampose: Fashion image-to-video synthesis via stable diffusion.
\newblock \emph{arXiv:2304.06025}, 2023.

\bibitem[Kawar et~al.(2023)Kawar, Zada, Lang, Tov, Chang, Dekel, Mosseri, and Irani]{imagic}
Bahjat Kawar, Shiran Zada, Oran Lang, Omer Tov, Huiwen Chang, Tali Dekel, Inbar Mosseri, and Michal Irani.
\newblock Imagic: Text-based real image editing with diffusion models.
\newblock In \emph{CVPR}, 2023.

\bibitem[Khachatryan et~al.(2023)Khachatryan, Movsisyan, Tadevosyan, Henschel, Wang, Navasardyan, and Shi]{Text2video-zero}
Levon Khachatryan, Andranik Movsisyan, Vahram Tadevosyan, Roberto Henschel, Zhangyang Wang, Shant Navasardyan, and Humphrey Shi.
\newblock Text2video-zero: Text-to-image diffusion models are zero-shot video generators.
\newblock \emph{arXiv:2303.13439}, 2023.

\bibitem[Kingma and Welling(2013)]{VAE}
Diederik~P Kingma and Max Welling.
\newblock Auto-encoding variational bayes.
\newblock \emph{arXiv:1312.6114}, 2013.

\bibitem[Li et~al.(2023)Li, Tucker, Snavely, and Holynski]{Generative-image-dynamics}
Zhengqi Li, Richard Tucker, Noah Snavely, and Aleksander Holynski.
\newblock Generative image dynamics.
\newblock \emph{arXiv:2309.07906}, 2023.

\bibitem[Liew et~al.(2023)Liew, Yan, Zhang, Xu, and Feng]{liew2023magicedit}
Jun~Hao Liew, Hanshu Yan, Jianfeng Zhang, Zhongcong Xu, and Jiashi Feng.
\newblock Magicedit: High-fidelity and temporally coherent video editing.
\newblock \emph{arXiv:2308.14749}, 2023.

\bibitem[Liu et~al.(2023{\natexlab{a}})Liu, Feng, Zhu, Zhang, Zheng, Liu, Zhao, Zhou, and Cao]{liu2023cones}
Zhiheng Liu, Ruili Feng, Kai Zhu, Yifei Zhang, Kecheng Zheng, Yu Liu, Deli Zhao, Jingren Zhou, and Yang Cao.
\newblock Cones: Concept neurons in diffusion models for customized generation.
\newblock \emph{arXiv:2303.05125}, 2023{\natexlab{a}}.

\bibitem[Liu et~al.(2023{\natexlab{b}})Liu, Zhang, Shen, Zheng, Zhu, Feng, Liu, Zhao, Zhou, and Cao]{cones2}
Zhiheng Liu, Yifei Zhang, Yujun Shen, Kecheng Zheng, Kai Zhu, Ruili Feng, Yu Liu, Deli Zhao, Jingren Zhou, and Yang Cao.
\newblock Cones 2: Customizable image synthesis with multiple subjects.
\newblock \emph{arXiv:2305.19327}, 2023{\natexlab{b}}.

\bibitem[Luan(2023)]{AnimateDiff-I2V}
Tyler Luan.
\newblock Animatediff-i2v.
\newblock \url{https://github.com/ykk648/AnimateDiff-I2V}, 2023.

\bibitem[Mahapatra and Kulkarni(2022)]{controllable-fluid-animation}
Aniruddha Mahapatra and Kuldeep Kulkarni.
\newblock Controllable animation of fluid elements in still images.
\newblock In \emph{CVPR}, 2022.

\bibitem[Meng et~al.(2021)Meng, He, Song, Song, Wu, Zhu, and Ermon]{meng2021sdedit}
Chenlin Meng, Yutong He, Yang Song, Jiaming Song, Jiajun Wu, Jun-Yan Zhu, and Stefano Ermon.
\newblock Sdedit: Guided image synthesis and editing with stochastic differential equations.
\newblock \emph{arXiv:2108.01073}, 2021.

\bibitem[Mou et~al.(2023)Mou, Wang, Xie, Zhang, Qi, Shan, and Qie]{T2i-adapter}
Chong Mou, Xintao Wang, Liangbin Xie, Jian Zhang, Zhongang Qi, Ying Shan, and Xiaohu Qie.
\newblock T2i-adapter: Learning adapters to dig out more controllable ability for text-to-image diffusion models.
\newblock \emph{arXiv:2302.08453}, 2023.

\bibitem[Okabe et~al.(2009)Okabe, Anjyo, Igarashi, and Seidel]{fluid-animating}
Makoto Okabe, Ken Anjyo, Takeo Igarashi, and Hans-Peter Seidel.
\newblock Animating pictures of fluid using video examples.
\newblock In \emph{Computer Graphics Forum}. Wiley Online Library, 2009.

\bibitem[Oquab et~al.(2023)Oquab, Darcet, Moutakanni, Vo, Szafraniec, Khalidov, Fernandez, Haziza, Massa, El-Nouby, et~al.]{dinov2}
Maxime Oquab, Timoth{\'e}e Darcet, Th{\'e}o Moutakanni, Huy Vo, Marc Szafraniec, Vasil Khalidov, Pierre Fernandez, Daniel Haziza, Francisco Massa, Alaaeldin El-Nouby, et~al.
\newblock Dinov2: Learning robust visual features without supervision.
\newblock \emph{arXiv:2304.07193}, 2023.

\bibitem[Podell et~al.(2023)Podell, English, Lacey, Blattmann, Dockhorn, M{\"u}ller, Penna, and Rombach]{podell2023sdxl}
Dustin Podell, Zion English, Kyle Lacey, Andreas Blattmann, Tim Dockhorn, Jonas M{\"u}ller, Joe Penna, and Robin Rombach.
\newblock Sdxl: Improving latent diffusion models for high-resolution image synthesis.
\newblock \emph{arXiv:2307.01952}, 2023.

\bibitem[Radford et~al.(2021)Radford, Kim, Hallacy, Ramesh, Goh, Agarwal, Sastry, Askell, Mishkin, Clark, et~al.]{CLIP}
Alec Radford, Jong~Wook Kim, Chris Hallacy, Aditya Ramesh, Gabriel Goh, Sandhini Agarwal, Girish Sastry, Amanda Askell, Pamela Mishkin, Jack Clark, et~al.
\newblock Learning transferable visual models from natural language supervision.
\newblock In \emph{Int. Conf. Mach. Learn.}, 2021.

\bibitem[reseachers(2023.10{\natexlab{a}})]{PikaLabs}
PikaLabs reseachers.
\newblock Pikalabs: An innovative text-to-video platform.
\newblock \url{https://www.pika.art/}, 2023.10{\natexlab{a}}.

\bibitem[reseachers(2023.10{\natexlab{b}})]{gen-2}
Runway reseachers.
\newblock Gen-2: The next step forward for generative ai.
\newblock \url{https://research.runwayml.com/gen2}, 2023.10{\natexlab{b}}.

\bibitem[Rombach et~al.(2022)Rombach, Blattmann, Lorenz, Esser, and Ommer]{ldm}
Robin Rombach, Andreas Blattmann, Dominik Lorenz, Patrick Esser, and Bj{\"o}rn Ommer.
\newblock High-resolution image synthesis with latent diffusion models.
\newblock In \emph{IEEE Conf. Comput. Vis. Pattern Recog.}, 2022.

\bibitem[Ruiz et~al.(2023)Ruiz, Li, Jampani, Pritch, Rubinstein, and Aberman]{dreambooth}
Nataniel Ruiz, Yuanzhen Li, Varun Jampani, Yael Pritch, Michael Rubinstein, and Kfir Aberman.
\newblock Dreambooth: Fine tuning text-to-image diffusion models for subject-driven generation.
\newblock In \emph{IEEE Conf. Comput. Vis. Pattern Recog.}, 2023.

\bibitem[Saharia et~al.(2022)Saharia, Chan, Saxena, Li, Whang, Denton, Ghasemipour, Gontijo~Lopes, Karagol~Ayan, Salimans, et~al.]{imagen}
Chitwan Saharia, William Chan, Saurabh Saxena, Lala Li, Jay Whang, Emily~L Denton, Kamyar Ghasemipour, Raphael Gontijo~Lopes, Burcu Karagol~Ayan, Tim Salimans, et~al.
\newblock Photorealistic text-to-image diffusion models with deep language understanding.
\newblock \emph{Adv. Neural Inform. Process. Syst.}, 2022.

\bibitem[Shalev and Wolf(2022)]{perturbed-masks-animation}
Yoav Shalev and Lior Wolf.
\newblock Image animation with perturbed masks.
\newblock In \emph{CVPR}, 2022.

\bibitem[Siarohin et~al.(2019)Siarohin, Lathuili{\`e}re, Tulyakov, Ricci, and Sebe]{first-order-animation}
Aliaksandr Siarohin, St{\'e}phane Lathuili{\`e}re, Sergey Tulyakov, Elisa Ricci, and Nicu Sebe.
\newblock First order motion model for image animation.
\newblock \emph{NeurIPS}, 2019.

\bibitem[Singer et~al.(2022)Singer, Polyak, Hayes, Yin, An, Zhang, Hu, Yang, Ashual, Gafni, et~al.]{Make-a-video}
Uriel Singer, Adam Polyak, Thomas Hayes, Xi Yin, Jie An, Songyang Zhang, Qiyuan Hu, Harry Yang, Oron Ashual, Oran Gafni, et~al.
\newblock Make-a-video: Text-to-video generation without text-video data.
\newblock \emph{arXiv:2209.14792}, 2022.

\bibitem[talesofai(2023)]{talesofai}
talesofai.
\newblock Animatediff talesofai.
\newblock \url{https://github.com/talesofai/AnimateDiff}, 2023.

\bibitem[Wang et~al.(2023{\natexlab{a}})Wang, Li, Lin, Lin, Yang, Zhang, Liu, and Wang]{wang2023disco}
Tan Wang, Linjie Li, Kevin Lin, Chung-Ching Lin, Zhengyuan Yang, Hanwang Zhang, Zicheng Liu, and Lijuan Wang.
\newblock Disco: Disentangled control for referring human dance generation in real world.
\newblock \emph{arXiv:2307.00040}, 2023{\natexlab{a}}.

\bibitem[Wang et~al.(2023{\natexlab{b}})Wang, Yuan, Zhang, Chen, Wang, Zhang, Shen, Zhao, and Zhou]{wang2023videocomposer}
Xiang Wang, Hangjie Yuan, Shiwei Zhang, Dayou Chen, Jiuniu Wang, Yingya Zhang, Yujun Shen, Deli Zhao, and Jingren Zhou.
\newblock Videocomposer: Compositional video synthesis with motion controllability.
\newblock \emph{NeurIPS}, 2023{\natexlab{b}}.

\bibitem[Wang et~al.(2004)Wang, Bovik, Sheikh, and Simoncelli]{SSIM}
Zhou Wang, Alan~C Bovik, Hamid~R Sheikh, and Eero~P Simoncelli.
\newblock Image quality assessment: from error visibility to structural similarity.
\newblock \emph{TIP}, 2004.

\bibitem[Wu et~al.(2023)Wu, Ge, Wang, Lei, Gu, Shi, Hsu, Shan, Qie, and Shou]{Tune-a-video}
Jay~Zhangjie Wu, Yixiao Ge, Xintao Wang, Stan~Weixian Lei, Yuchao Gu, Yufei Shi, Wynne Hsu, Ying Shan, Xiaohu Qie, and Mike~Zheng Shou.
\newblock Tune-a-video: One-shot tuning of image diffusion models for text-to-video generation.
\newblock In \emph{ICCV}, 2023.

\bibitem[Xue et~al.(2023)Xue, Song, Guo, Liu, Zong, Liu, and Luo]{xue2023raphael}
Zeyue Xue, Guanglu Song, Qiushan Guo, Boxiao Liu, Zhuofan Zong, Yu Liu, and Ping Luo.
\newblock Raphael: Text-to-image generation via large mixture of diffusion paths.
\newblock \emph{NeurIPS}, 2023.

\bibitem[Yin et~al.(2023)Yin, Wu, Yang, Wang, Wang, Ni, Yang, Li, Liu, Yang, et~al.]{yin2023nuwa}
Shengming Yin, Chenfei Wu, Huan Yang, Jianfeng Wang, Xiaodong Wang, Minheng Ni, Zhengyuan Yang, Linjie Li, Shuguang Liu, Fan Yang, et~al.
\newblock Nuwa-xl: Diffusion over diffusion for extremely long video generation.
\newblock \emph{arXiv:2303.12346}, 2023.

\bibitem[Zhang et~al.(2023)Zhang, Yan, Xu, Feng, and Liew]{zhang2023magicavatar}
Jianfeng Zhang, Hanshu Yan, Zhongcong Xu, Jiashi Feng, and Jun~Hao Liew.
\newblock Magicavatar: Multimodal avatar generation and animation.
\newblock \emph{arXiv:2308.14748}, 2023.

\bibitem[Zhang and Agrawala(2023)]{controlnet}
Lvmin Zhang and Maneesh Agrawala.
\newblock Adding conditional control to text-to-image diffusion models.
\newblock \emph{arXiv:2302.05543}, 2023.

\bibitem[Zhao and Zhang(2022)]{thin-plate-animation}
Jian Zhao and Hui Zhang.
\newblock Thin-plate spline motion model for image animation.
\newblock In \emph{CVPR}, 2022.

\bibitem[Zhao et~al.(2021)Zhao, Wu, and Guo]{sparse-to-dense-animation}
Ruiqi Zhao, Tianyi Wu, and Guodong Guo.
\newblock Sparse to dense motion transfer for face image animation.
\newblock In \emph{ICCV}, 2021.

\end{thebibliography}
